\documentclass[pdflatex,sn-nature]{sn-jnl}
\usepackage{graphicx}
\usepackage{amsmath}
\usepackage{booktabs}
\usepackage{xcolor}
\usepackage{placeins}
\usepackage{seqsplit}
\setcitestyle{super,comma,open={},close={}}

\unnumbered
\ExplSyntaxOn
\RenewDocumentCommand{\texttt}{m}{
  \group_begin:\ttfamily
  \int_compare:nNnTF {\tl_count:n {#1}} > {30}
    {\seqsplit{#1}} {#1}
  \group_end:
}
\ExplSyntaxOff
\newsavebox{\figurecaptionbox}
\newlength{\figureavailableheight}

\makeatletter
\def\addressfont{\reset@font\fontsize{8bp}{9.5bp}\selectfont\titraggedcenter}
\makeatother
\begin{document}
\title{tmol: a GPU-accelerated implementation of the Rosetta energy function for machine learning}
\author[1]{\fnm{Andrew} \sur{Leaver-Fay}}
\equalcont{Core Contributors.}
\author[2,3,4,5]{\fnm{Kieran} \sur{Didi}}
\equalcont{Core Contributors.}
\author[2,5]{\fnm{Jeff} \sur{Flatten}}
\equalcont{Core Contributors.}
\author[2,5]{\fnm{Alex} \sur{Ford}}
\author[6]{\fnm{Guangfeng} \sur{Zhou}}
\author[2,5,7]{\fnm{Nathaniel} \sur{Corley}}
\author[8,9]{\fnm{Hahnbeom} \sur{Park}}
\author[2,5,10]{\fnm{Maxwell W.} \sur{Kazman}}
\author[2,5,11]{\fnm{Rafael I.} \sur{Brent}}
\author[1]{\fnm{Joseph} \sur{Kleinhenz}}
\author[2,5]{\fnm{Henry} \sur{Solberg}}
\author[1]{\fnm{Andrew M.} \sur{Watkins}}
\author[12,13]{\fnm{Brian} \sur{Kuhlman}}
\author[2,5,14]{\fnm{David} \sur{Baker}}
\author*[2,5]{\fnm{Frank} \sur{DiMaio}}
\email{dimaio@uw.edu}
\affil[1]{\orgname{Prescient Design, Genentech, South San Francisco, CA, USA}}
\affil[2]{\orgname{Institute for Protein Design, University of Washington, Seattle, WA, USA}}
\affil[3]{\orgname{NVIDIA Corp., Santa Clara, USA}}
\affil[4]{\orgname{Department of Computer Science, University of Oxford, Oxford, UK}}
\affil[5]{\orgname{Department of Biochemistry, University of Washington, Seattle, WA, USA}}
\affil[6]{\orgname{Center for Advanced Therapeutics, Wistar Institute, Philidelphia, PA, USA}}
\affil[7]{\orgname{Department of Bioengineering, University of Washington, Seattle, WA, USA}}
\affil[8]{\orgname{Biomedical Research Division, Korea Institute of Science and Technology, Seoul, Republic of Korea}}
\affil[9]{\orgname{KIST-SKKU Brain Research Center, SKKU Institute for Convergence, Sungkyunkwan University, Suwon, Republic of Korea}}
\affil[10]{\orgname{Graduate Program in Biological Physics, Structure and Design, University of Washington, Seattle, WA, USA}}
\affil[11]{\orgname{Graduate Program in Molecular Engineering, University of Washington, Seattle, WA, USA}}
\affil[12]{\orgname{Department of Biochemistry and Biophysics, School of Medicine, University of North Carolina at Chapel Hill, Chapel Hill, NC, USA}}
\affil[13]{\orgname{Lineberger Comprehensive Cancer Center, University of North Carolina at Chapel Hill, Chapel Hill, NC, USA}}
\affil[14]{\orgname{Howard Hughes Medical Institute, University of Washington, Seattle, WA, USA}}

\abstract{Physics-inspired forcefields complement generative machine learning models, showing better generalization beyond the training distribution.  However, these forcefields are typically slow to evaluate, and their derivatives are difficult to integrate into ML frameworks.  We introduce tmol, a GPU-native PyTorch library implementing the state-of-the-art RosettaGenFF forcefield. tmol supports efficient batched differentiable scoring and whole-structure refinement with up to 362-fold throughput increase (batched). Used as a reward in generative protein design, tmol increases hydrogen bonds and water bridges at designed interfaces.}
\maketitle

\section{Introduction}

Generative machine learning (ML) models have led to dramatic improvements in our ability to predict and design biomolecular structures\cite{abramson2024,krishna2024,watson2023}. However, such models generalize poorly to underrepresented chemistry: ligand-docking accuracy falls for unfamiliar pockets and ligands, where generated models often show poor stereochemistry\cite{buttenschoen2024,xu2026,masters2025}; and RNA structure prediction degrades sharply on novel folds\cite{bernard2025}. These limitations make physical forcefields appealing as priors, identifying physically implausible conformations that generative models may miss.  Unfortunately, integrating such forcefields has been difficult, as they are often too slow to incorporate into an ML pipeline, and incompatible interfaces makes efforts to do so laborious.  GPU-accelerated forcefields\cite{eastman2024,doerr2021} address part of this, accelerating molecular-mechanics simulations, but the Rosetta energy function and its sampling algorithms, including discrete side-chain optimization and torsion-space minimization, have had no differentiable, GPU-native counterpart.

Here we introduce tmol, an open-source PyTorch library that addresses both of these issues by reimplementing the RosettaGenFF forcefield\cite{park2016,park2021} as highly efficient batched tensor operations. Building on the successful Rosetta software package\cite{leaverfay2011,alford2017}, we also implement a set of ``core algorithms,'' including whole-structure and residue-pair scoring, discrete side-chain optimization (packing), Cartesian and torsion-space minimization \cite{conway2014}, and a general-purpose refinement protocol (relax) \cite{maguire2021}. Scoring, packing and minimization are all native PyTorch operations returning coordinate or torsion derivatives through autograd\cite{paszke2019}, so energies and refinement may compose directly into ML model training and inference.

\section{Results}

Figure~\ref{fig:architecture} outlines how tmol is organized. A molecular system is divided into blocks (amino-acid residues or ligands), and energies are computed between pairs of neighboring blocks. Each block pair is assigned to its own GPU thread block, a cooperative thread array (CTA) containing 32 threads (Fig.~\ref{fig:architecture}b). These threads copy the coordinates and parameters for both blocks into shared memory (a fast scratch space), then evaluate all atom pairs between the blocks (up to $32\times32$), with each thread handling a subset. Each thread's computed energy is then summed across the CTA to give one energy per score term for the block pair; each atom pair's derivative contribution is added directly to a per-atom gradient tensor. Blocks with more than 32 atoms are processed in several such tiles. For minimization over torsion angles, tmol also converts between internal coordinates (bond lengths, angles and torsions) and Cartesian coordinates on the GPU, using a parallel algorithm called a segmented scan (see Methods).

Within this framework, all the terms of RosettaGenFF were reimplemented.  Small changes addressed three issues (details in Methods): inconsistent count-pair weighting, double-counted hydroxyl torsions, and representation of tabulated potentials.  We also implemented a widely used Rosetta refinement protocol, \textit{FastRelax} \cite{maguire2021}, with an additional change to improve rotamer recovery metrics (see Methods).  Following testing on a diverse set of 1000 high-resolution protein structures, tmol reproduced PyRosetta score terms closely (Pearson $r\geq0.97$ for every term and 0.997 for the total; Fig.~\ref{fig:architecture}d).  Further, tmol is significantly faster than Rosetta: over a range of protein sizes, tmol achieves 233--362x speedup for score-plus-gradient calculations, and 38--102x speedup for torsion-space relax relative to single-threaded PyRosetta (Fig.~\ref{fig:architecture}c).  This improvement is despite the fact that relax in tmol considers ~60\% more rotamers than PyRosetta.

One challenge in using Rosetta and other biomolecular forcefields is that configuring novel chemistry is often only ``semi-automated,'' requiring users to manually check protonation, charge assignment, and atom and bond types.  Proper handling of non-standard residues and metals is quite challenging.  tmol includes a fast, fully automatic input-processing pipeline that requires no user input: ligands, modified amino acids, D-amino acids, modified nucleotides and most metallocomplexes are parameterized directly from the deposited chemistry.  A new metal-coordination term (see Methods) as well as DNA and RNA torsion potentials---derived with the scheme of Yanover and Bradley\cite{yanover2011} from DNA and RNA structures, respectively---properly evaluate internal energetics of such complexes.

Evaluating tmol on previously published benchmarks used to optimize and assess energy-function quality\cite{park2016,park2021,glasscock2025,zhou2024}, we observe similar performance across a wide range of modalities, including protein-monomer, protein--protein, and protein--DNA decoy discrimination, side-chain-rotamer recovery, ligand docking, and stability $\Delta\Delta G$ (Fig.~\ref{fig:applications}b).  All differences fall within run-to-run variation except rotamer recovery, where tmol gains by keeping near-native rotamers that Rosetta's bump check discards.  Ligand docking highlights the strength of the automatic pipeline: generating PyRosetta parameters required several minutes of processing and manual inspection of input models, while tmol took about 1 second to generate parameters for each ligand with no user input.

To demonstrate integration with machine learning frameworks, we used tmol as a reward in beam search with the protein-design model RFD4-Proteina \cite{corley2026rfd4proteina}, scoring candidate binders without gradients (Online Methods). Across 24 protein--protein targets, a hydrogen-bond reward raised the share of designs with at least five interface residue pairs making hydrogen bonds from 16\% to 66\% (53\% for the best 32 of 128 unguided designs chosen on the same quantity), and a \texttt{lk\_bridge} ``bridging-water'' reward raised the share with strong water bridging from 17\% to 68\% (55\% for best-of-$k$) (Fig.~\ref{fig:applications}c--e). Forward folding \textit{in silico} success, assessed with OpenFold3\cite{openfold3}, rose from 14.5\% to 22.9\% (hydrogen-bond reward), matching but not exceeding best-of-$k$ selection (25.5\%); search designs also had larger interfaces and slightly more clashes.

\section{Discussion}

tmol provides a framework that allows generative ML methods, physics-based forcefields, and physics-guided simulations to all naturally interoperate. It is our hope that tmol will enable the rapid development of ML models that incorporate energetics tightly into their networks.  Many problems in biomolecular structure prediction \cite{bernard2025,skrinjar2025} show poor generalization due to limited data: obtaining high-resolution structures is costly and labor-intensive. For these cases, we believe tmol represents a strong ``biophysical prior'' that should prove valuable in improving model generalizability.

\backmatter

\section*{Declarations}


\textbf{Competing interests}\quad {The authors declare the following competing interests: A.L.-F., J.K. and A.M.W. are employees of Prescient Design, Genentech, a member of the Roche Group. K.D. is an employee of NVIDIA. tmol is distributed under the open-source BSD 3-Clause license, and no patents have been filed on the work described. The remaining authors declare no competing interests.}

\textbf{Author contributions}\quad A.L.-F., A.F., D.B. and F.D. conceived the project. A.L.-F., A.F., F.D., K.D., J.F. and H.S. developed the tmol code. G.Z. and H.P. contributed to the ligand and noncanonical-residue processing pipeline, K.D, M.W.K., R.I.B. and N.C. contributed to the AtomWorks integration, and J.K. contributed to supporting method development. F.D., A.L.-F., K.D. and J.F. curated the data and performed the analyses. F.D., D.B., B.K. and A.M.W. supervised the work. A.L.-F., K.D. and F.D. wrote the first draft of the manuscript. All authors edited the manuscript and approved the final version.


\textbf{Code availability}\quad tmol is available at \url{https://github.com/RosettaCommons/tmol}; documentation is at \url{https://rosettacommons.github.io/tmol/latest/}. 

\textbf{Ethics approval}\quad Not applicable.

\clearpage
{\setlength{\bibsep}{8pt}

}
\clearpage
\setcounter{figure}{0}\renewcommand{\thefigure}{\arabic{figure}}%
  \newgeometry{left=12mm,right=12mm,top=10mm,bottom=18mm,bindingoffset=0mm}%
  \thispagestyle{headings}%
  \begingroup
  \setlength{\parskip}{0pt}%
  \sbox{\figurecaptionbox}{\begin{minipage}{\textwidth}%
    \figurecaptionfont\textbf{Figure 1 | Batched molecular modeling and performance.} \textbf{a}, tmol's shared molecular representation and modeling operations. \textbf{b}, Pairs of interacting molecular blocks (residues) map to GPU thread blocks (cooperative thread arrays, CTAs). Each block-pair's 32 threads reuse atom data in shared memory to calculate energies and coordinate gradients. \textbf{c}, Throughput of score-plus-gradient and FastRelax as a function of structure size for PyRosetta (one CPU core), tmol on one or 32 CPU threads, and tmol on one GPU at batch sizes B of 1, 10 and 100. Points are medians; brackets give the fold change of tmol (B100) over PyRosetta at 800 residues. \textbf{d}, Agreement with Rosetta on 1,000 high-resolution protein chains: Pearson correlation (left; 95\% bootstrap intervals) and tmol minus Rosetta energy per residue (right; mean $\pm$ s.d. over chains) for each score term. Shading marks terms affected by the deliberate changes to count-pair weighting, interpolation and double counting.%
  \end{minipage}}%
  \setlength{\figureavailableheight}{\dimexpr\textheight-\ht\figurecaptionbox-\dp\figurecaptionbox-5mm\relax}%
  \refstepcounter{figure}\label{fig:architecture}%
  \noindent\begin{minipage}[t]{\textwidth}\centering
    \includegraphics[width=\textwidth,height=\figureavailableheight,keepaspectratio]{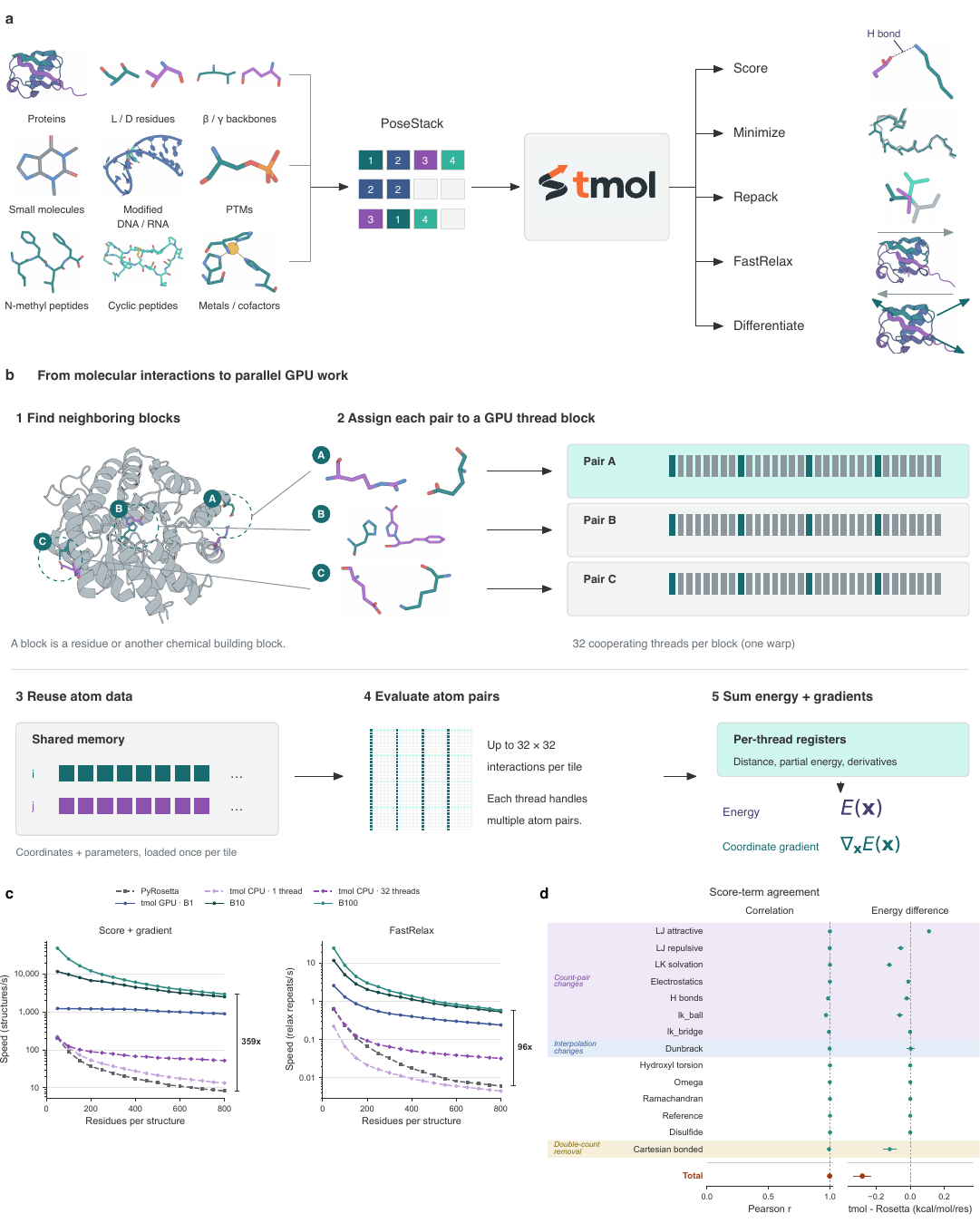}%
    \par\vspace{2mm}\noindent\usebox{\figurecaptionbox}%
  \end{minipage}\par
  \endgroup
  \restoregeometry

  \newgeometry{left=12mm,right=12mm,top=10mm,bottom=18mm,bindingoffset=0mm}%
  \thispagestyle{headings}%
  \begingroup
  \setlength{\parskip}{0pt}%
  \sbox{\figurecaptionbox}{\begin{minipage}{\textwidth}%
    \figurecaptionfont\textbf{Figure 2 | Molecular energetics and tmol-guided protein design.} \textbf{a}, Score-function components; all 26 energy terms appear in Online Methods (Table~\ref{tab:score-channels}). \textbf{b}, Energy-function benchmark summaries; higher is better. \textbf{c}, Reward-guided beam search with tmol interface energies for 24 protein--protein targets. Bars show the percentage of designs with at least five hydrogen-bonded binder--target residue pairs (tmol \texttt{hbond} $\leq-0.3$; left) or with strong water bridging (summed tmol \texttt{lk\_bridge} energy $\leq-0.15$; right) for unguided sampling, the best 32 of 128 unguided designs chosen on the plotted quantity (best-of-$k$), and search with a hydrogen-bond or \texttt{lk\_bridge} reward; error bars are 95\% bootstrap intervals over targets. \textbf{d}, Example search designs (hydrogen-bond reward) with 10 and 9 interface hydrogen bonds, all shown; dashed lines join hydrogen-bonded residue pairs. \textbf{e}, Example search designs (\texttt{lk\_bridge} reward) with 6 and 5 water bridges. Spheres are implied waters, each placed 2.7--3.1~\AA\ from both polar atoms of an \texttt{lk\_bridge} residue pair ($\leq-0.03$) without steric clash.%
  \end{minipage}}%
  \setlength{\figureavailableheight}{\dimexpr\textheight-\ht\figurecaptionbox-\dp\figurecaptionbox-5mm\relax}%
  \refstepcounter{figure}\label{fig:applications}%
  \noindent\begin{minipage}[t]{\textwidth}\centering
    \includegraphics[width=\textwidth,height=\figureavailableheight,keepaspectratio]{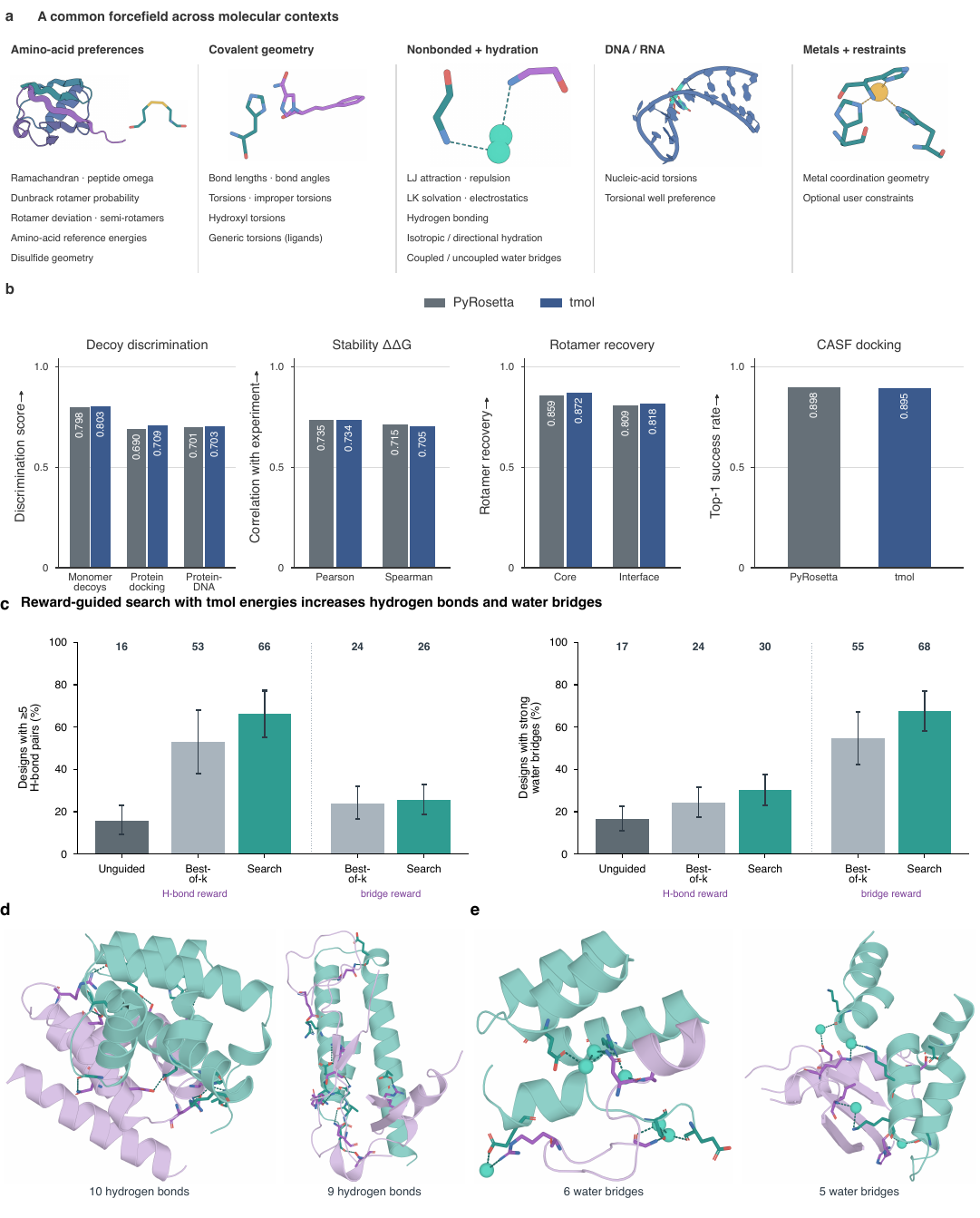}%
    \par\vspace{2mm}\noindent\usebox{\figurecaptionbox}%
  \end{minipage}\par
  \endgroup
  \restoregeometry

\clearpage
\section*{Online Methods}

\subsection*{Software versions and reproducibility}
Unless otherwise specified, all comparisons use tmol version 0.2.0, PyTorch 2.14.1, CUDA 13.0 and AtomWorks\cite{corley2025atomworks} 3.0.0; Rosetta comparisons use PyRosetta\cite{chaudhury2010} 2026.3+releasequarterly.

\subsection*{PoseStack: a GPU-native molecular representation}
Rosetta3 represents molecular topology, coordinates and parameters through an object-oriented molecular model\cite{leaverfay2011}; independent trajectories can be parallelized across multiple CPUs.  GPU execution, however, requires enough concurrent work to overcome launch overhead, and needs contiguous inputs addressable by every participating thread.  Consequently, tmol combines independent systems into a single \texttt{PoseStack}. A \texttt{PoseStack} represents its principal coordinates as a dense tensor of shape \texttt{[n\_poses, max\_n\_atoms, 3]}.  In this representation, all model parameters are represented as per-\emph{block} (see below) padded tensors.  Members of a \texttt{PoseStack} may contain multiple unrelated structures with differing residue counts, sequences and chemistries; these structures do not interact with each other.  Rather, the role of a \texttt{PoseStack} is to keep the GPU saturated.

The basic unit of calculation in tmol is the \emph{block}. Each block represents a set of atoms with a corresponding \emph{block type}, which defines its atoms, intra-block bonds, chemical parameters, and a set of connection points through which it can bond to other blocks. Nothing in this interface assumes a polymer, so the same representation covers polymer residues, small-molecule ligands, and fragments of larger chemical components that have been split across several connected blocks. Blocks play the role in tmol that \texttt{Residue}s play in Rosetta: energies are evaluated over blocks and between pairs of blocks.

A \texttt{PoseStack} represents a batch of poses as padded tensors. For each pose it stores 1) a \texttt{PackedBlockTypes} object describing an ordered list of block types, 2) the block-type identity for each block in the \texttt{PoseStack} as an index into the \texttt{PackedBlockTypes} object, 3) a set of coordinates for all atoms in all poses, 4) a per-block offset into the coordinate tensor, and 5) for each connection point on each block, the block and connection point it is bonded to. Poses with fewer blocks or atoms are padded to the largest pose in the batch. (\texttt{PoseStack}s also contain a \texttt{ConstraintSet}, a \texttt{PDBInfo} object for convenient input and output, and the number of chemical bonds separating all pairs of inter-block connection points.)

\texttt{PackedBlockTypes} holds an ordered list of block types. The order is fixed at construction, so a \texttt{PoseStack} can refer to a block type by index. Per-type properties such as atom counts, bonds, torsions and connection atoms are stored as padded tensors with one row per block type. \texttt{PackedBlockTypes} also serves as a persistent cache for the tensors that score terms and algorithms derive from the block types in order to perform their calculations efficiently, an idiom we call \emph{annotation}. Rather than adding term-specific fields to the chemical database, each term computes its inputs for each block type as CPU arrays, annotates them onto the block type, and then concatenates these per-block annotations, in \texttt{PackedBlockTypes} order, into device-resident tensors annotated onto the \texttt{PackedBlockTypes}. This annotation cost is thus only paid once per set of block types rather than once per pose or once per scoring evaluation. Terms reuse an existing annotation when one is already present, and shared base classes let several terms reuse common annotations such as the bond graph, bond separations, and atom types.

A \texttt{PoseStack}'s chemistry and connectivity are fixed for its lifetime; only its coordinates change. Minimization therefore updates a stack in place, whereas any change to topology, batch membership, or sequence (as in packing or design) produces a new stack. Because a \texttt{PoseStack}'s chemistry is immutable, stacks can be shallow-copied cheaply.

\subsection*{Parameterizing new chemistry}

The default database for tmol contains precomputed parameters for: the \textbf{canonical L-amino acids}, including protonation and tautomer variants (\texttt{HIS\_D} and \texttt{HIS\_POS}, and deprotonated forms of Cys, His, Lys and Tyr for metal coordination) and disulfide-bonded cysteine; \textbf{D-amino acids}, generated as mirror images of the L residues that read the L backbone tables at negated $\phi/\psi$; \textbf{DNA and RNA nucleotides}; \textbf{water}; and \textbf{termini variants} for both polymer types. It also includes 29 \textbf{metal-ion} residue types, one per combination of ion and coordination geometry, covering 12 elements, and 17 \textbf{multi-metal cofactors} such as iron-sulfur clusters. The default score function is beta\_nov2016 extended with nucleic-acid torsion, generic torsion (gen\_bonded) and metal-coordination terms.

Any residue other than these basic types is parameterized automatically. Ligands can be supplied as a SMILES string, a MOL2/SDF file or a CIF file, or detected directly in a Biotite\cite{kunzmann2018} structure as residues absent from the active database. Bond orders \emph{must come from the input}: tmol does not infer bonds from geometry. Prepared residues are added to a new \texttt{ParameterDatabase} (extending the input one) and can be saved to an editable \texttt{.tmol} file for reuse.

Parameterization proceeds in three steps. First, the residue is protonated at the requested pH by Dimorphite-DL\cite{dimorphite2019} (with a few added rules). Second, a reference structure is generated from distance geometry using custom PyTorch code, followed by a short force-field minimization using MMFF94\cite{halgren1996} in Open Babel\cite{oboyle2011}. MMFF94 partial charges are also assigned from this computation. Third, atoms are typed with a port of Rosetta's generic-potential classifier\cite{park2021}, and parameters are derived: bond lengths and angles are restrained about their values in the reference structure, and torsions are scored by the generic gen\_bonded term.

Covalently linked residues require additional handling. A noncanonical residue within a polymer chain is capped at its polymer connections, parameterized as above, and then uncapped. The backbone atoms of an $\alpha$-amino acid are retyped to the canonical protein types so that the protein-specific backbone terms apply, while its side chain keeps generic types. Residues with other backbones keep generic types throughout. A residue attached through a side chain, such as a glycan or covalent ligand, is joined to its partner by a generated patch that adds a connection at the attachment site.

\subsection*{Chemistry beyond canonical residues}

\paragraph{Noncanonical residues.} Noncanonical $\alpha$-amino acids are given hybrid parameters: backbone atoms receive the canonical protein atom types, while side-chain atoms keep the generic types assigned during preparation. Their backbone is scored by the protein backbone torsion terms (Ramachandran and omega ($\omega$)), using the tables of the most similar canonical amino acid, or alanine if none matches; side-chain torsions are scored by \texttt{gen\_bonded}. All other noncanonical amino acids, including $\beta$- and $\gamma$-amino acids and N-methylated residues, keep generic types throughout and are scored entirely by \texttt{gen\_bonded}.

For packing, the side chain of every noncanonical amino acid, $\alpha$ or not, is matched against the canonical side chains, and the rotamer library of the most similar one is borrowed. The borrowed library is used only to generate rotamers and never contributes to the score. Side chains with no match, and any torsions beyond those the borrowed library covers, are sampled by enumerating a fixed set of values for each rotatable bond.

\paragraph{Metal coordination.} The coordination geometry around a metal is inferred from the input structure. When a pose is built, candidate donor atoms near each metal are collected, and the coordination geometries allowed for that metal are fitted to the donor directions to find the best match. Two things follow. First, each coordinating residue is patched to bond to the metal and, if needed, switched to a deprotonated variant that can donate. Second, virtual atoms are built around the metal at ideal coordination sites, oriented by the fit. A metal-coordination energy term then restrains each coordinating atom to its site, holding it at the ideal metal--donor distance (median distances from high-resolution structures) and along the site direction. Multi-metal cofactors such as iron-sulfur clusters are handled the same way, with some sites filled by the cofactor's own atoms.

\paragraph{Ligand fragmentation.}  While tmol can only subdivide interaction energies down to the block level, for large ligands it is desirable to further subdivide energies into ligand substructures.  tmol provides a mechanism for this: annotating a ligand's atoms with \texttt{tmol\_fragment\_id} splits it into separate blocks which are each scored separately, allowing ligand energies to be reported at finer granularity.  The ligand is parameterized as a whole before it is split, so the total score is unchanged by fragmentation.

\subsection*{Score functions and scoring modes}
A score function in tmol is a weighted list of energy terms. Scoring a \texttt{PoseStack} proceeds in two phases. In the first, \emph{rendering}, each term annotates the block types and the pose with whatever data it needs; these annotations depend only on the chemistry of the pose, not its coordinates. The result is a scoring module bound to that \texttt{PoseStack}. In the second phase, the module passes coordinates and annotations to the scoring kernels and returns the weighted sum of the term energies. Because rendering does not depend on coordinates, a single rendered module can score any number of conformations of the same \texttt{PoseStack}, as in minimization.

A score function can be rendered in one of three modes, all built on the same block-level calculations. \textbf{Whole-pose scoring} returns one weighted energy per pose, optionally broken down by score term. \textbf{Block-pair scoring} returns one energy per pair of blocks in each pose. This gives per-residue and interface energies directly. Combined with automatic differentiation, it also allows individual interactions to be weighted differently. Block-pair scoring carries a small overhead in runtime and memory. \textbf{Rotamer scoring} is used during packing: it scores every candidate rotamer against every other and returns a sparse rotamer-pair energy tensor, which is traversed during simulated annealing.

The API includes a function \texttt{beta2016\_score\_function}, which returns a \texttt{ScoreFunction} object with weights and terms analogous to Rosetta's \texttt{beta\_nov2016\_cart}, with additional terms for ligand and nucleic-acid torsions and for metal coordination. It is also possible to construct a \texttt{ScoreFunction} from a weights file by using \texttt{ScoreFunction.from\_sfxn\_file(...)} and passing in the path to that file.

\subsection*{Efficient, autograd-compatible CPU, CUDA, and Metal (MPS) scoring kernels}

Individual scoreterm kernels were designed with three goals: efficiency on CPU and on both NVIDIA (CUDA) and Apple-silicon (Metal/MPS) GPUs, a single shared implementation across devices wherever possible, and full compatibility with PyTorch automatic differentiation. The sections below describe how the code is shared between devices, how the principal non-bonded kernels are organized on the GPU, and how derivatives are exposed to PyTorch.

\paragraph{Unified CPU, CUDA and Metal execution}
The greatest risk in maintaining separate CPU and GPU versions of a function is that the two (or more!) implementations diverge: a bug is fixed in one but not the other, or a feature is added to only one, and the two begin computing different values. tmol avoids this through heavy use of templates. The lowest-level operations are written once, in a device-agnostic way, as C++ lambda functions, and the same code is then handed to one of several device-specific launchers. On the CPU, the lambda is called in a \texttt{for} loop over all $N$ inputs; on the GPU, it is passed to a moderngpu \cite{baxter2016} \texttt{transform} operation, which wraps it in a kernel and launches that kernel with $N$ threads. Device-specific code is thereby confined to a small execution layer whose two versions can be checked for equivalence by inspection. This layer, the \texttt{DeviceOperations} class, provides kernel launches, scans, and reductions, with its CPU and CUDA versions in separate files.

Each kernel follows the same execution ``diamond'' (Fig.~\ref{fig:diamond}). A C++ entry point calls a method of a \texttt{struct} templated on the device holding the tensors. The CPU, CUDA and Metal instantiations of that \texttt{struct} live in separate \texttt{.cpu.cpp}, \texttt{.cuda.cu} and \texttt{.mps.cpp} files, which all include the same generically written \texttt{.impl.hh} file along with the corresponding \texttt{DeviceOperations} implementation: the paths diverge by device and then merge again in shared code. Eigen's tensor module\cite{eigen} is used as a shared abstraction on CPU and CUDA; code shared with Metal uses a small portable math layer that also compiles as Metal Shading Language.

\paragraph{Apple-silicon (MPS) kernels}
Metal threadgroups on Apple GPUs execute in 32-wide SIMD groups, just like CUDA CTAs execute in 32-thread warps, so the block-pair tiling used on CUDA (see \emph{Block-pair GPU kernels} below) runs on Metal largely unchanged (with threadgroup memory in place of CUDA shared memory). The two GPUs are used differently, however. Apple GPUs are slower than CUDA devices, but moving data between host and GPU is cheap. Due to this, on CUDA, all scoreterms and operations stay on the device; on Metal, coordinates and data stay in host tensors and only a handful of compute-intensive operations are offloaded to the device, with results copied back.  Offloaded terms include \texttt{fa\_ljatr}, \texttt{fa\_ljrep}, \texttt{fa\_lk} and \texttt{fa\_elec} in whole-pose scoring, and additionally  \texttt{lk\_ball} and \texttt{hbond} in rotamer-pair scoring for packing.  Parallel simulated annealing is also carried out on-device. The remaining terms are evaluated on the CPU concurrently.  While only limited testing was carried out, on a base M5 processor, we see ~3.5-fold speedup using this approach (comparing all-CPU tmol to all-CPU+device).

All device paths are nevertheless tested separately, because reduction order and floating-point arithmetic remain device dependent.

\paragraph{Block-pair GPU kernels}
The principal non-bonded kernels operate on pairs of blocks. tmol places a bounding sphere of radius $r_i$ around each block and selects block pairs with $d_{ij}<r_i+r_j+r_{\mathrm{cut}}$, where $r_{\mathrm{cut}}$ is the interaction reach of the term. Atom-pair evaluation then applies each term's own cutoff and bond-separation exclusions (e.g., the Lennard--Jones cutoff is 6.0~\AA).

On the GPU, a grid of 32-thread cooperative thread arrays (CTAs), each a single warp (on Metal, a 32-thread threadgroup), strides through the list of selected block pairs, with each CTA handling one pair at a time. A CTA stages the coordinates and parameters of up to 32 atoms from each block in shared memory. On a full tile, thread $t$ evaluates the flattened atom-pair indices $t+32k$ for $k=0,\ldots,31$, so the tile's 1,024 interactions are spread across 32 threads while reusing the staged atom data. Larger blocks are processed as successive tiles. Each thread's energy contributions are reduced within the CTA and added to the pose total, and Cartesian derivatives are accumulated atomically at the affected coordinates.

This tiling scheme is written with templates and shared by all four atom-pair energy terms: Lennard--Jones/Lazaridis--Karplus\cite{lazaridis1999}, LK-ball, electrostatics and hydrogen bonding. New atom-pair terms can adopt it with little additional code.

Accumulating energies by block pair also improves numerical stability. Early versions of tmol added every atom-pair contribution atomically into a single global total. This proved unstable, with energies varying by $\sim 10^{-3}$ between runs on identical inputs because of nondeterministic thread execution order: adding many small numbers to a large one loses their significant figures \cite{kahan1965}. Having each warp first compute a block-pair subtotal over roughly 400 atom-pair interactions (about 20 atoms per block) and then accumulating these larger subtotals is substantially more stable, and also reduces global-memory traffic. Because the unit of work is a block pair, whole-pose, block-pair and rotamer scoring all reuse the same atom-pair functions.

\paragraph{Automatic differentiation}
Each compiled scoring operator provides an explicit backward function to PyTorch, so a weighted energy $E(\mathbf{x})$ can be differentiated with respect to the Cartesian coordinates $\mathbf{x}$ and combined with other differentiable objectives. This supports Cartesian minimization, gradient-based guidance of generative models, and training losses built from tmol energies. For efficiency, many kernels in whole-pose scoring compute derivatives during the forward pass when gradients are required and cache them for the backward pass, avoiding a second traversal of the interacting atom pairs. Block-pair scoring does not do this, which accounts for its small overhead.

\subsection*{Parallel internal-coordinate kinematics}

One key feature of Rosetta is internal-coordinate (or ``torsion-space'') refinement. Treating bond lengths and angles as ideal allows optimization in a lower-dimensional space, which gives both faster convergence and less susceptibility to getting stuck in local minima. Molecular dynamics and molecular mechanics packages typically sample in Cartesian space, where the fastest bond and angle vibrations limit the usable step size; Rosetta instead treats these degrees of freedom as ideal and samples in torsion space. It does this by constructing a kinematic tree \cite{wang2007}, so that changes to torsional degrees of freedom can be propagated into new Cartesian coordinates by a recursive traversal of the tree. A significant challenge for tmol was that GPUs are poorly suited to recursion.

To solve this, we turn to the GPU algorithm {\em scan}\ \cite{harris2007}, which is frequently used to compute prefix sums ({\em e.g.}\ index offsets). In $O(n \log n)$ parallel steps, scan can compute
\[ B_j = A_0 \otimes A_1 \otimes \cdots \otimes A_{j-1} \]
for all $j < n$ (with $B_0$ the identity), and crucially, this works for any binary, associative operator $\otimes$.  If we represent the transform from the coordinate frame of a parent atom to that of its child as a homogeneous transform, then the operation to compute that child's frame is matrix multiplication, and matrix multiplication is binary and associative. Thus, if we break Rosetta's kinematic tree into a set of paths, we can repeatedly use scan to compute the coordinates for all atoms, and using a variation on scan, {\em segmented scan}, can process multiple paths at the same ``depth'' at once. We call this algorithm {\em generational segmented scan}, as each generation reaches a new set of branches in the tree.


For user control over the kinematic tree, tmol represents block-level connectivity with a batched \texttt{FoldForest} (the batched equivalent of Rosetta's \texttt{FoldTree}) and internally expands this to an atom-level \texttt{KinForest}. \texttt{FoldForest} represents three types of connections between blocks: {\em polymer}\ edges describe connected block ranges, {\em jump}\ edges connect arbitrary blocks with six rigid-body degrees of freedom, and {\em root jump}\ edges connect each molecular tree to a virtual root at the origin, allowing rigid-body orientation of the entire system, necessary for support of anisotropic energy terms. Each pose in the \texttt{PoseStack} must have at least one root jump edge, but may contain arbitrarily many.

The process of converting from the \texttt{FoldForest} to the \texttt{KinForest} is complex. We wanted assembly of this data structure to involve the CPU as little as possible, though it is most readily expressed recursively. We factor the portions of a full \texttt{KinForest} into the several scan-path segments that each block contributes. The number of different variations of scan-path-segment assemblies for each block type is given by the product of the number of input-connection points and the number of output-connection points, and these variations can all be precomputed; we call these stencils. With enough careful construction of offsets, we can assemble the stencils into a full \texttt{KinForest}. The first step, though, involves a post-order, depth-first traversal of the edges in the \texttt{FoldForest} on the CPU that determines the ``generational delay'' for each edge -- the first generation that the blocks of that edge will be traversed -- converting the partial order of the edges into a total order. From there forward, everything can be performed on the GPU.

The same forward algorithm that starts at the root and travels down to the leaves to compute coordinates from internal DOFs also works backwards to sum the $f$ and $g$ derivative vectors \cite{abe1984rapid} (called ``$f_1$'' and ``$f_2$'' in Rosetta and tmol) from the leaves up toward the root. tmol can therefore minimize with respect to internal degrees of freedom, just as Rosetta does. The $g$ vectors ($f_2$) are the coordinate gradients $\partial E/\partial \mathbf{r}_i$ returned by the backward pass, and the $f$ vectors ($f_1$) follow from them as $f_i = g_i \times \mathbf{r}_i$, where $\mathbf{r}_i$ is the position of atom $i$.  This shortcut lets the backward pass compute only the $g$ vectors; the $f$ vectors are computed only when using internal-DOF minimization.

tmol uses the work-efficient scan in moderngpu\cite{baxter2016}, so forward kinematics performs work that is $O(n \log n)$ in the number of atoms, with a parallel depth set by the number of nested branches in the kinematic tree rather than by chain length; a single protein chain requires five generations regardless of its size.

\subsection*{Rotamer packing and simulated annealing}

Packing is a basic Rosetta operation that finds side-chain conformations optimizing the energy for a given (fixed) backbone.  The packing layer in tmol: a) constructs candidate rotamers from pluggable samplers; b) computes a sparse rotamer-pair energy graph using the rotamer scoring mode of the score function; and c) runs a simulated anneal trajectory to find the rotamer assignment minimizing total energy.  A \texttt{PackerTask} controls allowed identities and positions. Fixed-sequence repacking and compatible sequence design are implemented. Compiled CPU and CUDA annealers consume the same interaction-graph representation but use device-specific search strategies. Unlike the scoring kernels, which must agree atom pair for atom pair, annealing is a stochastic search: on the GPU many short parallel trajectories reach comparable energies faster than one long trajectory, so the CUDA annealer performs more total work than the CPU annealer while taking less time. The CPU annealer is a faithful reproduction of Rosetta's standard annealing algorithm \cite{kuhlman2000}: the temperature is lowered from 100 to 0.3 over 20 outer cycles, each of five inner steps per rotamer, followed by a greedy quench from the best assignment found.

The CUDA annealer maps one 32-thread warp to one trajectory. Threads cooperatively score a proposed rotamer substitution; one lane performs the Metropolis decision and broadcasts it across the warp. The default search begins with 500 independently initialized high-temperature trajectories per pose. A short greedy quench provides comparable ranking scores, the best 25\% are retained, and each retained state seeds ten lower-temperature trajectories, yielding 1,250 candidates. After a second short quench and 25\% retention, 312 trajectories proceed to a full greedy quench. Counter-based Philox\cite{salmon2011} random-number streams make the schedule reproducible from the PyTorch CUDA generator state.  We do not benchmark the CUDA annealer against Rosetta's directly. However, two results indicate its equivalence: the energy-function benchmarks above were run on GPU and invoke the packer throughout, and FastRelax energy changes agreed between tmol and PyRosetta within run-to-run variation at every size tested.

\subsection*{Minimization and FastRelax}
Rosetta uses a very specific minimization formula: L-BFGS with an Armijo line search\cite{liu1989,armijo1966}.  While the use of any torch-compatible minimizer is possible, we have ported this exact minimization strategy to torch. The tmol implementation minimizes every pose in a stack simultaneously but independently: each pose keeps its own L-BFGS history and line search, and converges and stops on its own, so minimizing a batch gives the same result as minimizing each pose separately. The L-BFGS update uses the compact matrix representation of the inverse Hessian \cite{byrd1994}, so it runs in parallel on the GPU.  One small change was made from Rosetta's default convergence criteria: rather than requiring both gradient magnitude and improvement to be small, tmol requires five consecutive steps with small improvement with no restrictions on gradient magnitude.  This shortens some minimization trajectories without increasing energies.

Cartesian and internal coordinate (Torsion-space) minimization are both supported, and both use the same rendered scoring module and kernels. Cartesian minimization optimizes atom coordinates directly. Torsion-space minimization optimizes the selected torsional and rigid-body degrees of freedom $\theta$: each energy evaluation first passes through the kinematic layer, which rebuilds Cartesian coordinates from $\theta$ by generational segmented scan, and the resulting coordinates are then scored. Because the kinematic layer is itself an autograd operation, whose backward pass accumulates the $f_1$/$f_2$ vectors described above, PyTorch automatically composes $\partial E/\partial \mathbf{x}$ with $\partial \mathbf{x}/\partial \theta$ to give $\partial E/\partial \theta$. No minimization-specific derivative code is needed.

Finally, \texttt{fast\_relax} is Rosetta's default ``refinement'' protocol. \texttt{fast\_relax} alternates packing and minimization while ramping repulsive and optional constraint weights. Its four default packing/minimization repulsive fractions are 0.040/0.051, 0.265/0.280, 0.559/0.581 and 1.000/1.000, reproducing the Monomer Relax schedule from Maguire et al.\ \cite{maguire2021}. It lives in the \texttt{tmol.relax} module as a short Python protocol built directly on the packing and minimization layers described above, so its schedule can be changed by argument and new relax-style protocols can be written in the same way.

\subsection*{Performance benchmarks}

We compared tmol with PyRosetta (release 2026.3+releasequarterly) on two tasks using the beta\_nov16\_cart energy function. The first is a single \textit{torsion-space} score and gradient evaluation. The second is \textit{torsion-space} \texttt{fast\_relax} with five repeats.  Inputs were compact subsets of chain A of PDB 1DP0, from 50 to 800 residues. Hydrogens were placed once by Rosetta, and tmol read these coordinates without further
optimization, with both programs started from identical structures. PyRosetta minimized with lbfgs\_armijo\_nonmonotone, while tmol used its GPU-batched L-BFGS implementation.  Both were capped at 200 minimization steps.  Although the two protocols are not identical, relax energy changes from the two programs agreed within 5\% at every size tested (tmol on average 2\% smaller in magnitude).

All timings were run on a 32-core AMD EPYC 9355 processor and H200 NVL GPUs, using PyTorch 2.14.1 with CUDA 13.0. PyRosetta ran on one CPU core. tmol ran on one or 32 CPU cores, and for GPU calculations, one GPU with four host cores.  GPU calculations were run with batches of 1, 10 or 100 copies of each structure. Score and gradient times are the median of 20 calls after three warm-up calls; FastRelax times are the median of three runs after one warm-up
run. We report throughput as structures processed per second, counting each FastRelax repeat as one step.  Note that due to the removal of the ``bump check'' in packing (see below), tmol considers about 60\% more rotamers in relax than PyRosetta.

\subsection*{Differences from Rosetta's implementation}
tmol reproduces the functional forms and parameters of RosettaGenFF\cite{park2021} (beta\_nov2016 with gen\_bonded), with deliberate changes in two areas. First, count-pair weighting is applied uniformly. Rosetta computes bond separation only between directly bonded residues, so pairs such as C$_i$ and N$_{i+2}$, four bonds apart, receive full weight. It also uses a three-bond rather than four-bond crossover across non-peptide inter-residue bonds. tmol computes shortest bond paths through intermediate blocks and applies a consistent rule for all nonbonded terms: pairs fewer than four bonds apart are excluded, and pairs exactly four bonds apart are scaled by 0.2.

Second, we removed redundant terms and simplified tabulated potentials. Rosetta restrains the Ser, Thr and Tyr hydroxyl hydrogens twice, in both cart\_bonded and the hydroxyl-torsion term; tmol restrains them once.  Rosetta restrains peptide planarity with impropers centered on N and C, but one of its restraints (O$_{i}$--C$_{i}$--N$_{i+1}$--H$_{i+1}$) is a proper torsion about the peptide bond that duplicates the $\omega$ term; tmol uses impropers only. The Ramachandran, $\omega$, \texttt{p\_aa\_pp}, and Dunbrack\cite{shapovalov2011} tables use the same cubic-spline interpolation as Rosetta, except for the Dunbrack chi means and standard deviations, which Rosetta interpolates linearly, while tmol uses cubic-splines. For Ramachandran, $\omega$, and Dunbrack, these match Rosetta exactly at grid points but differ slightly between them. The \texttt{p\_aa\_pp} term, however, is interpolated at positions $5^\circ$ off the 0-centered grid points of the other three terms. We thus interpolated the \texttt{p\_aa\_pp} values at the 0-centers and then fit the splines to pass through the interpolated values.

Finally, in packing, we made one small change from Rosetta. When building backbone-dependent rotamers, Rosetta by default applies a ``bump check'' that discards any candidate rotamer whose repulsive interaction with the surrounding ``background'' (backbone and non-sampled side chains) exceeds a fixed threshold, before rotamer pair energies are computed. This removes about 40\% of candidate rotamers, but it sometimes discards rotamers that would be part of the lowest-energy solution. Because rotamer pair energies are cheaper to compute on the GPU, tmol disables
the bump check by default and considers every rotamer during annealing---about 60\% more than PyRosetta. The check is still available as an option on the packer task.  Energy-function weights were not re-optimized after any of these changes; benchmark performance is unchanged within error (Fig.~\ref{fig:applications}b).

\subsection*{Agreement with Rosetta}

To compare tmol's energies with Rosetta's, we scored a set of high-resolution protein structures with both programs. Starting from a PISCES \cite{PISCES} culled list (X-ray structures at $\leq$1.5~\AA{} resolution, R $\leq$ 0.25, $\leq$25\% pairwise sequence identity, no chain breaks; October 2026), we kept chains of 50--500 residues, additionally requiring R-free $\leq$ 0.25, and took the 1,000 most recently
deposited chains that extracted cleanly, keeping the first model and highest-occupancy alternate locations for each chain and removing waters, ligands and metals. Hydrogens were added once in PyRosetta. The resulting files were scored without modification (no hydrogen re-optimization) by both Rosetta (beta\_nov16\_cart) and tmol, summing related Rosetta terms into the matching tmol term (e.g., \texttt{fa\_atr} and \texttt{fa\_intra\_atr\_xover4} for \texttt{fa\_ljatr}). Energies are weighted and expressed per residue. For each term we report the Pearson correlation across chains, with a 95\% confidence interval from 2,000 bootstrap samples, and the mean and standard deviation of the per-residue difference between the two programs.

\subsection*{Reward-guided design with tmol interface energies}
RFD4-Proteina (release checkpoints) designed binders against 24 protein--protein benchmark targets. Reward-guided search used beam search: 16 replicates per target branched into four candidates at denoising steps 30 and 60 (of 60), kept the two highest-reward candidates per replicate and returned 32 designs. Candidates were completed by a rollout and scored without gradients using tmol block-pair energies between binder and target residues. The reward was the number of residue pairs with \texttt{hbond} $\leq-0.3$ (hydrogen-bond reward) or \texttt{lk\_bridge} $\leq-0.05$ (bridge reward), minus $0.5\times$ or $10\times$ the summed pair energy, respectively. Controls were unguided sampling (32 designs) and the best 32 of 128 unguided designs on the plotted quantity (best-of-$k$). All designs were rescored identically with tmol. Reported percentages use criteria fixed independently of the rewards: a design counts as hydrogen bonded when at least five binder--target residue pairs have \texttt{hbond} $\leq-0.3$, and as strongly water bridged when its summed \texttt{lk\_bridge} energy over binder--target pairs is $\leq-0.15$. Percentages are per-target means with bootstrap intervals over targets. Designs were refolded with OpenFold3\cite{openfold3} and judged with the protein-binder criterion (ptm $\geq80$, minimum ipAE $\leq1.5$, binder RMSD $\leq2.5$~\AA).

\subsection*{Energy-function benchmarks}
We compared tmol and PyRosetta on benchmarks previously used to develop and validate Rosetta energy functions. From Park et al.\ \cite{park2016}, we used three tasks. \emph{Monomer decoy discrimination} (61 proteins) and \emph{protein--protein docking decoy discrimination} (65 complexes) ask whether the energy function ranks near-native structures below non-native decoys, measured with the Boltzmann-weighted discrimination score of Park et al.\ \cite{park2016}. \emph{Stability $\Delta\Delta G$} (about 1,200 point mutations) compares predicted and experimental changes in folding free energy by Pearson and Spearman correlation. \emph{Side-chain-rotamer recovery} follows the density-based protocol of Conway and DiMaio\cite{conway2016}, on a subset of their benchmark (167 core and 88 interface targets).  Side chains are rebuilt by the packer from the backbone alone, and each predicted side chain is compared with an experimental electron density map: a side chain counts as ``recovered'' when its density correlation falls no more than 0.13 below that of the native conformation and is at least 0.70. Alanine and glycine are excluded, as are positions with maximum side chain B-factor $\geq30$, alternate conformations, possible crystal contacts, or poorly resolved density. Both tmol and PyRosetta predictions are scored by the same PyRosetta density calculation. From Glasscock et al.\ \cite{glasscock2025}, \emph{protein--DNA decoy discrimination} (24 complexes) uses the same discrimination score. Following Zhou et al.\ \cite{zhou2024}, we used the \emph{CASF-2016} protein--ligand set\cite{casf2016} (285 complexes), reporting docking power as the fraction of complexes whose top-ranked pose lies within 2~\AA\ of the crystal pose, and scoring power as the correlation of predicted with experimental binding affinity. In all tasks, both engines relaxed or minimized the same inputs with the same movable residues, using a single-repeat Cartesian FastRelax or, for CASF, a short restrained minimization.

\subsection*{Score-function components and molecular illustrations}
Figure~\ref{fig:applications}a groups the implemented score channels by their physical roles. Table~\ref{tab:score-channels} gives the complete mapping to code identifiers and documents the representative structures.

\clearpage
{\setlength{\bibsep}{5pt}

}
\clearpage
\setcounter{figure}{0}\renewcommand{\thefigure}{S\arabic{figure}}%
\begin{figure}[p]
\centering
\includegraphics[width=0.8\textwidth]{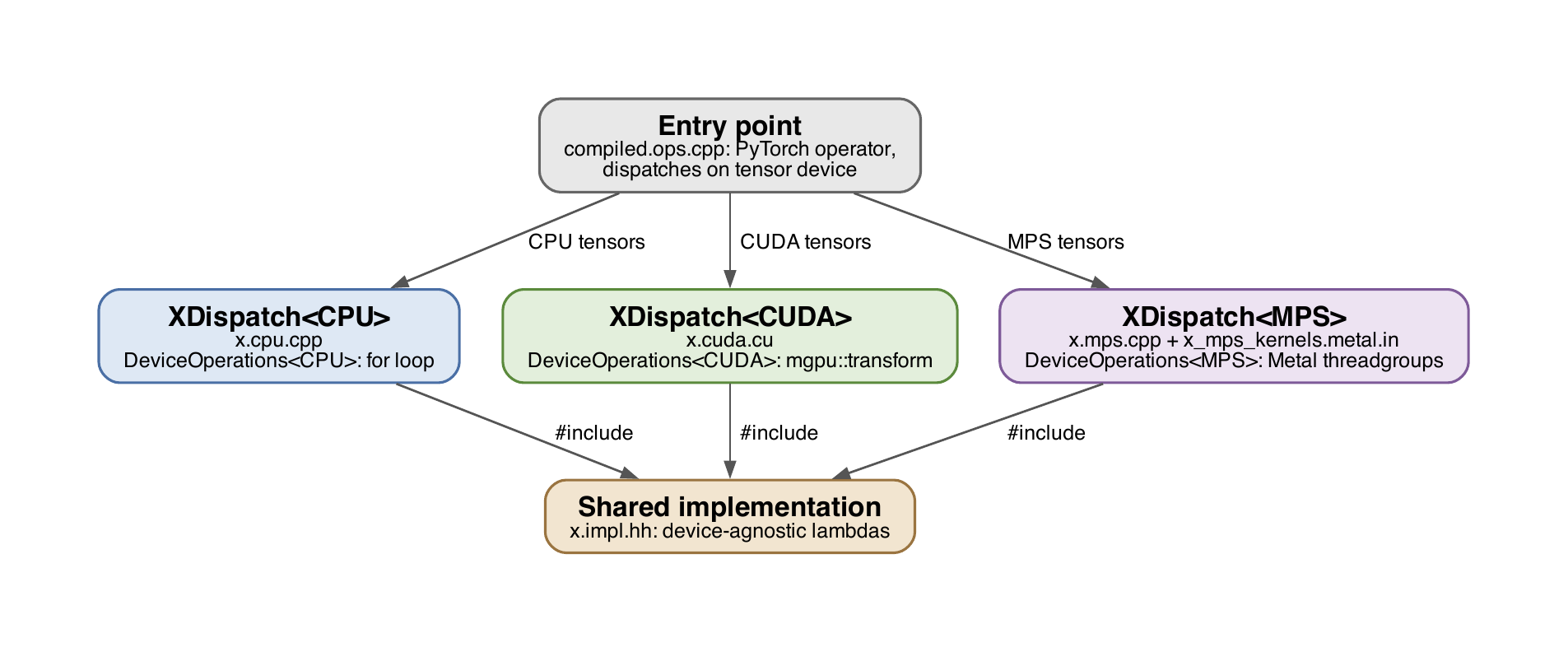}
\caption{Device dispatch in tmol. A single device-agnostic implementation is compiled separately for CPU, CUDA and Apple Metal (MPS); only the launch layer differs between devices.}
\label{fig:diamond}
\end{figure}
\clearpage
\setcounter{table}{0}\renewcommand{\thetable}{S\arabic{table}}%
\begin{table}[p]
\centering
\caption{Score channels represented in Figure~\ref{fig:applications}a. All 26 identifiers in \texttt{tmol/score/\_score\_types.py} are listed; \texttt{n\_score\_types} is a count sentinel, not an energy term. The inspected checkout and benchmark source agree on this list.}\label{tab:score-channels}
\small
\begin{tabular}{@{}p{0.55\textwidth}p{0.41\textwidth}@{}}
\toprule
Readable term & Code identifier \\
\midrule
\multicolumn{2}{@{}l}{\textbf{Amino-acid preferences}} \\
Ramachandran backbone preference & \texttt{rama} \\
Peptide omega torsion & \texttt{omega} \\
Dunbrack rotamer probability & \texttt{dunbrack\_rot} \\
Dunbrack rotamer deviation & \texttt{dunbrack\_rotdev} \\
Dunbrack semi-rotameric term & \texttt{dunbrack\_semirot} \\
Amino-acid reference energy & \texttt{ref} \\
Disulfide geometry & \texttt{disulfide} \\
\addlinespace[3pt]
\multicolumn{2}{@{}l}{\textbf{Covalent geometry}} \\
\multicolumn{2}{@{}l}{Cartesian bonded (\texttt{cart\_bonded})} \\
Bond lengths & \texttt{cart\_lengths} \\
Bond angles & \texttt{cart\_angles} \\
Bond torsions & \texttt{cart\_torsions} \\
Improper torsions & \texttt{cart\_impropers} \\
Hydroxyl torsions & \texttt{cart\_hxltorsions} \\
Generic bonded (\texttt{gen\_bonded}) & \texttt{gen\_torsions} \\
\addlinespace[3pt]
\multicolumn{2}{@{}l}{\textbf{Nonbonded and hydration}} \\
Lennard--Jones attraction & \texttt{fa\_ljatr} \\
Lennard--Jones repulsion & \texttt{fa\_ljrep} \\
Lazaridis--Karplus solvation & \texttt{fa\_lk} \\
Electrostatics & \texttt{fa\_elec} \\
Hydrogen bonding & \texttt{hbond} \\
Directional implicit hydration & \texttt{lk\_ball} \\
Isotropic implicit hydration & \texttt{lk\_ball\_iso} \\
Coupled water-bridge term & \texttt{lk\_bridge} \\
Uncoupled water-bridge term & \texttt{lk\_bridge\_uncpl} \\
\addlinespace[3pt]
\multicolumn{2}{@{}l}{\textbf{DNA / RNA}} \\
Nucleic-acid torsions & \texttt{na\_torsion} \\
Nucleic-acid torsional wells & \texttt{na\_torsion\_well} \\
\addlinespace[3pt]
\multicolumn{2}{@{}l}{\textbf{Metals and restraints}} \\
Metal coordination geometry & \texttt{metal\_coordination} \\
User-defined constraints & \texttt{constraint} \\
\bottomrule
\end{tabular}
\par\medskip
\noindent\begin{minipage}{0.98\textwidth}\footnotesize\raggedright
Figure~\ref{fig:architecture}d groups \texttt{cart\_lengths}, \texttt{cart\_angles}, \texttt{cart\_torsions} and \texttt{cart\_impropers} as \texttt{cart\_bonded}; \texttt{cart\_hxltorsions} is plotted separately as hydroxyl torsion. Its \texttt{gen\_bonded} comparison uses \texttt{gen\_torsions}. These grouped comparison labels are not additional \texttt{ScoreType} identifiers. 

\end{minipage}
\end{table}


\begin{thebibliography}{10}
\expandafter\ifx\csname url\endcsname\relax
  \def\url#1{\burl{#1}}\fi
\expandafter\ifx\csname urlprefix\endcsname\relax\def\urlprefix{URL }\fi
\providecommand{\bibinfo}[2]{#2}
\providecommand{\eprint}[2][]{\url{#2}}
\providecommand{\doi}[1]{\url{https://doi.org/#1}}
\bibcommenthead

\bibitem{abramson2024}
\bibinfo{author}{Abramson, J.} \emph{et~al.}
\newblock \bibinfo{title}{Accurate structure prediction of biomolecular
  interactions with {AlphaFold 3}}.
\newblock \emph{\bibinfo{journal}{Nature}} \textbf{\bibinfo{volume}{630}},
  \bibinfo{pages}{493--500} (\bibinfo{year}{2024}).


\bibitem{krishna2024}
\bibinfo{author}{Krishna, R.} \emph{et~al.}
\newblock \bibinfo{title}{Generalized biomolecular modeling and design with
  {RoseTTAFold All-Atom}}.
\newblock \emph{\bibinfo{journal}{Science}} \textbf{\bibinfo{volume}{384}},
  \bibinfo{pages}{eadl2528} (\bibinfo{year}{2024}).


\bibitem{watson2023}
\bibinfo{author}{Watson, J.~L.} \emph{et~al.}
\newblock \bibinfo{title}{De novo design of protein structure and function with
  {RFdiffusion}}.
\newblock \emph{\bibinfo{journal}{Nature}} \textbf{\bibinfo{volume}{620}},
  \bibinfo{pages}{1089--1100} (\bibinfo{year}{2023}).


\bibitem{buttenschoen2024}
\bibinfo{author}{Buttenschoen, M.}, \bibinfo{author}{Morris, G.~M.} \&
  \bibinfo{author}{Deane, C.~M.}
\newblock \bibinfo{title}{{PoseBusters}: {AI}-based docking methods fail to
  generate physically valid poses or generalise to novel sequences}.
\newblock \emph{\bibinfo{journal}{Chem. Sci.}} \textbf{\bibinfo{volume}{15}},
  \bibinfo{pages}{3130--3139} (\bibinfo{year}{2024}).


\bibitem{xu2026}
\bibinfo{author}{Xu, S.} \emph{et~al.}
\newblock \bibinfo{title}{Benchmarking all-atom biomolecular structure
  prediction with {FoldBench}}.
\newblock \emph{\bibinfo{journal}{Nat. Commun.}} \textbf{\bibinfo{volume}{17}},
  \bibinfo{pages}{442} (\bibinfo{year}{2026}).


\bibitem{masters2025}
\bibinfo{author}{Masters, M.~R.}, \bibinfo{author}{Mahmoud, A.~H.} \&
  \bibinfo{author}{Lill, M.~A.}
\newblock \bibinfo{title}{Investigating whether deep learning models for
  co-folding learn the physics of protein--ligand interactions}.
\newblock \emph{\bibinfo{journal}{Nat. Commun.}} \textbf{\bibinfo{volume}{16}},
  \bibinfo{pages}{8854} (\bibinfo{year}{2025}).


\bibitem{bernard2025}
\bibinfo{author}{Bernard, C.}, \bibinfo{author}{Postic, G.},
  \bibinfo{author}{Ghannay, S.} \& \bibinfo{author}{Tahi, F.}
\newblock \bibinfo{title}{Has {AlphaFold3} achieved success for {RNA}?}
\newblock \emph{\bibinfo{journal}{Acta Crystallogr. D Struct. Biol.}}
  \textbf{\bibinfo{volume}{81}}, \bibinfo{pages}{49--62}
  (\bibinfo{year}{2025}).


\bibitem{leaverfay2011}
\bibinfo{author}{Leaver-Fay, A.} \emph{et~al.}
\newblock \bibinfo{title}{{ROSETTA3}: an object-oriented software suite for the
  simulation and design of macromolecules}.
\newblock \emph{\bibinfo{journal}{Methods Enzymol.}}
  \textbf{\bibinfo{volume}{487}}, \bibinfo{pages}{545--574}
  (\bibinfo{year}{2011}).


\bibitem{alford2017}
\bibinfo{author}{Alford, R.~F.} \emph{et~al.}
\newblock \bibinfo{title}{The {Rosetta} all-atom energy function for
  macromolecular modeling and design}.
\newblock \emph{\bibinfo{journal}{J. Chem. Theory Comput.}}
  \textbf{\bibinfo{volume}{13}}, \bibinfo{pages}{3031--3048}
  (\bibinfo{year}{2017}).


\bibitem{eastman2024}
\bibinfo{author}{Eastman, P.} \emph{et~al.}
\newblock \bibinfo{title}{{OpenMM 8}: molecular dynamics simulation with
  machine learning potentials}.
\newblock \emph{\bibinfo{journal}{J. Phys. Chem. B}}
  \textbf{\bibinfo{volume}{128}}, \bibinfo{pages}{109--116}
  (\bibinfo{year}{2024}).


\bibitem{doerr2021}
\bibinfo{author}{Doerr, S.} \emph{et~al.}
\newblock \bibinfo{title}{{TorchMD}: a deep learning framework for molecular
  simulations}.
\newblock \emph{\bibinfo{journal}{J. Chem. Theory Comput.}}
  \textbf{\bibinfo{volume}{17}}, \bibinfo{pages}{2355--2363}
  (\bibinfo{year}{2021}).


\bibitem{park2016}
\bibinfo{author}{Park, H.} \emph{et~al.}
\newblock \bibinfo{title}{Simultaneous optimization of biomolecular energy
  functions on features from small molecules and macromolecules}.
\newblock \emph{\bibinfo{journal}{J. Chem. Theory Comput.}}
  \textbf{\bibinfo{volume}{12}}, \bibinfo{pages}{6201--6212}
  (\bibinfo{year}{2016}).


\bibitem{park2021}
\bibinfo{author}{Park, H.}, \bibinfo{author}{Zhou, G.}, \bibinfo{author}{Baek,
  M.}, \bibinfo{author}{Baker, D.} \& \bibinfo{author}{DiMaio, F.}
\newblock \bibinfo{title}{{Force field optimization guided by small molecule
  crystal lattice data enables consistent sub-angstrom protein--ligand
  docking}}.
\newblock \emph{\bibinfo{journal}{J. Chem. Theory Comput.}}
  \textbf{\bibinfo{volume}{17}}, \bibinfo{pages}{2000--2010}
  (\bibinfo{year}{2021}).


\bibitem{conway2014}
\bibinfo{author}{Conway, P.}, \bibinfo{author}{Tyka, M.~D.},
  \bibinfo{author}{DiMaio, F.}, \bibinfo{author}{Konerding, D.~E.} \&
  \bibinfo{author}{Baker, D.}
\newblock \bibinfo{title}{Relaxation of backbone bond geometry improves protein
  energy landscape modeling}.
\newblock \emph{\bibinfo{journal}{Protein Sci.}} \textbf{\bibinfo{volume}{23}},
  \bibinfo{pages}{47--55} (\bibinfo{year}{2014}).


\bibitem{maguire2021}
\bibinfo{author}{Maguire, J.~B.} \emph{et~al.}
\newblock \bibinfo{title}{Perturbing the energy landscape for improved packing
  during computational protein design}.
\newblock \emph{\bibinfo{journal}{Proteins}} \textbf{\bibinfo{volume}{89}},
  \bibinfo{pages}{436--449} (\bibinfo{year}{2021}).


\bibitem{paszke2019}
\bibinfo{author}{Paszke, A.} \emph{et~al.} \bibinfo{title}{{PyTorch}: an
  imperative style, high-performance deep learning library}.
\newblock in \emph{\bibinfo{booktitle}{Advances in Neural Information
  Processing Systems}}, Vol.~\bibinfo{volume}{32}, \bibinfo{pages}{8024--8035}
  (\bibinfo{year}{2019}).


\bibitem{yanover2011}
\bibinfo{author}{Yanover, C.} \& \bibinfo{author}{Bradley, P.}
\newblock \bibinfo{title}{Extensive protein and {DNA} backbone sampling
  improves structure-based specificity prediction for {C2H2} zinc fingers}.
\newblock \emph{\bibinfo{journal}{Nucleic Acids Res.}}
  \textbf{\bibinfo{volume}{39}}, \bibinfo{pages}{4564--4576}
  (\bibinfo{year}{2011}).


\bibitem{glasscock2025}
\bibinfo{author}{Glasscock, C.~J.} \emph{et~al.}
\newblock \bibinfo{title}{{Computational design of sequence-specific
  DNA-binding proteins}}.
\newblock \emph{\bibinfo{journal}{Nat. Struct. Mol. Biol.}}
  \textbf{\bibinfo{volume}{32}}, \bibinfo{pages}{2252--2261}
  (\bibinfo{year}{2025}).


\bibitem{zhou2024}
\bibinfo{author}{Zhou, G.} \emph{et~al.}
\newblock \bibinfo{title}{{An artificial intelligence accelerated virtual
  screening platform for drug discovery}}.
\newblock \emph{\bibinfo{journal}{Nat. Commun.}} \textbf{\bibinfo{volume}{15}},
  \bibinfo{pages}{7761} (\bibinfo{year}{2024}).


\bibitem{corley2026rfd4proteina}
\bibinfo{author}{Corley, N.} \emph{et~al.}
\newblock \bibinfo{title}{Biomolecular design through latent reasoning with
  {RFD4-Proteina}} (\bibinfo{year}{2026}).
\newblock \bibinfo{note}{Manuscript in preparation}.


\bibitem{openfold3}
\bibinfo{author}{{The OpenFold3 Team}}.
\newblock \bibinfo{title}{{OpenFold3-preview}} (\bibinfo{year}{2025}).
\newblock \urlprefix\url{https://github.com/aqlaboratory/openfold-3}.
\newblock \bibinfo{note}{Version 0.4.2}.


\bibitem{skrinjar2025}
\bibinfo{author}{{\v S}krinjar, P.}, \bibinfo{author}{Eberhardt, J.},
  \bibinfo{author}{Durairaj, J.} \& \bibinfo{author}{Schwede, T.}
\newblock \bibinfo{title}{Have protein--ligand co-folding methods moved beyond
  memorisation?}
\newblock \emph{\bibinfo{journal}{bioRxiv}}  (\bibinfo{year}{2025}).

\end{thebibliography}

\begin{thebibliography}{10}
\expandafter\ifx\csname url\endcsname\relax
  \def\url#1{\burl{#1}}\fi
\expandafter\ifx\csname urlprefix\endcsname\relax\def\urlprefix{URL }\fi
\providecommand{\bibinfo}[2]{#2}
\providecommand{\eprint}[2][]{\url{#2}}
\providecommand{\doi}[1]{\url{https://doi.org/#1}}
\bibcommenthead

\setcounter{NAT@ctr}{22}
\bibitem{corley2025atomworks}
\bibinfo{author}{Corley, N.} \emph{et~al.}
\newblock \bibinfo{title}{Accelerating biomolecular modeling with {AtomWorks}
  and {RF3}}.
\newblock \emph{\bibinfo{journal}{bioRxiv}}  (\bibinfo{year}{2025}).


\bibitem{chaudhury2010}
\bibinfo{author}{Chaudhury, S.}, \bibinfo{author}{Lyskov, S.} \&
  \bibinfo{author}{Gray, J.~J.}
\newblock \bibinfo{title}{{PyRosetta}: a script-based interface for
  implementing molecular modeling algorithms using {Rosetta}}.
\newblock \emph{\bibinfo{journal}{Bioinformatics}}
  \textbf{\bibinfo{volume}{26}}, \bibinfo{pages}{689--691}
  (\bibinfo{year}{2010}).


\bibitem{kunzmann2018}
\bibinfo{author}{Kunzmann, P.} \& \bibinfo{author}{Hamacher, K.}
\newblock \bibinfo{title}{{Biotite}: a unifying open source computational
  biology framework in {Python}}.
\newblock \emph{\bibinfo{journal}{BMC Bioinformatics}}
  \textbf{\bibinfo{volume}{19}}, \bibinfo{pages}{346} (\bibinfo{year}{2018}).


\bibitem{dimorphite2019}
\bibinfo{author}{Ropp, P.~J.}, \bibinfo{author}{Kaminsky, J.~C.},
  \bibinfo{author}{Yablonski, S.} \& \bibinfo{author}{Durrant, J.~D.}
\newblock \bibinfo{title}{{Dimorphite-DL: an open-source program for
  enumerating the ionization states of drug-like small molecules}}.
\newblock \emph{\bibinfo{journal}{J. Cheminform.}}
  \textbf{\bibinfo{volume}{11}}, \bibinfo{pages}{14} (\bibinfo{year}{2019}).


\bibitem{halgren1996}
\bibinfo{author}{Halgren, T.~A.}
\newblock \bibinfo{title}{{Merck molecular force field. I. Basis, form, scope,
  parameterization, and performance of MMFF94}}.
\newblock \emph{\bibinfo{journal}{J. Comput. Chem.}}
  \textbf{\bibinfo{volume}{17}}, \bibinfo{pages}{490--519}
  (\bibinfo{year}{1996}).


\bibitem{oboyle2011}
\bibinfo{author}{O'Boyle, N.~M.} \emph{et~al.}
\newblock \bibinfo{title}{{Open Babel: An open chemical toolbox}}.
\newblock \emph{\bibinfo{journal}{J. Cheminform.}}
  \textbf{\bibinfo{volume}{3}}, \bibinfo{pages}{33} (\bibinfo{year}{2011}).


\bibitem{baxter2016}
\bibinfo{author}{Baxter, S.}
\newblock \bibinfo{title}{moderngpu 2.0} (\bibinfo{year}{2016}).
\newblock \urlprefix\url{https://github.com/moderngpu/moderngpu/wiki}.


\bibitem{eigen}
\bibinfo{author}{{Eigen developers}}.
\newblock \emph{\bibinfo{title}{{Eigen Tensors}}}.
\newblock
  \urlprefix\url{https://eigen.tuxfamily.org/dox/unsupported/eigen_tensors.html}.
\newblock \bibinfo{note}{Accessed 10 October 2026}.


\bibitem{lazaridis1999}
\bibinfo{author}{Lazaridis, T.} \& \bibinfo{author}{Karplus, M.}
\newblock \bibinfo{title}{Effective energy function for proteins in solution}.
\newblock \emph{\bibinfo{journal}{Proteins}} \textbf{\bibinfo{volume}{35}},
  \bibinfo{pages}{133--152} (\bibinfo{year}{1999}).


\bibitem{kahan1965}
\bibinfo{author}{Kahan, W.}
\newblock \bibinfo{title}{Pracniques: Further remarks on reducing truncation
  errors}.
\newblock \emph{\bibinfo{journal}{Commun. ACM}} \textbf{\bibinfo{volume}{8}},
  \bibinfo{pages}{40} (\bibinfo{year}{1965}).


\bibitem{wang2007}
\bibinfo{author}{Wang, C.}, \bibinfo{author}{Bradley, P.} \&
  \bibinfo{author}{Baker, D.}
\newblock \bibinfo{title}{Protein--protein docking with backbone flexibility}.
\newblock \emph{\bibinfo{journal}{J. Mol. Biol.}}
  \textbf{\bibinfo{volume}{373}}, \bibinfo{pages}{503--519}
  (\bibinfo{year}{2007}).


\bibitem{harris2007}
\bibinfo{author}{Harris, M.}, \bibinfo{author}{Sengupta, S.} \&
  \bibinfo{author}{Owens, J.~D.}
\newblock \bibinfo{title}{Parallel prefix sum (scan) with {CUDA}} in
  \emph{\bibinfo{booktitle}{GPU Gems 3}} (ed. \bibinfo{editor}{Nguyen, H.})
  \bibinfo{pages}{851--876} (\bibinfo{publisher}{Addison-Wesley},
  \bibinfo{year}{2007}).


\bibitem{abe1984rapid}
\bibinfo{author}{Abe, H.}, \bibinfo{author}{Braun, W.},
  \bibinfo{author}{Noguti, T.} \& \bibinfo{author}{G{\={o}}, N.}
\newblock \bibinfo{title}{Rapid calculation of first and second derivatives of
  conformational energy with respect to dihedral angles for proteins: General
  recurrent equations}.
\newblock \emph{\bibinfo{journal}{Comput. Chem.}} \textbf{\bibinfo{volume}{8}},
  \bibinfo{pages}{239--247} (\bibinfo{year}{1984}).


\bibitem{kuhlman2000}
\bibinfo{author}{Kuhlman, B.} \& \bibinfo{author}{Baker, D.}
\newblock \bibinfo{title}{Native protein sequences are close to optimal for
  their structures}.
\newblock \emph{\bibinfo{journal}{Proc. Natl Acad. Sci. USA}}
  \textbf{\bibinfo{volume}{97}}, \bibinfo{pages}{10383--10388}
  (\bibinfo{year}{2000}).


\bibitem{salmon2011}
\bibinfo{author}{Salmon, J.~K.}, \bibinfo{author}{Moraes, M.~A.},
  \bibinfo{author}{Dror, R.~O.} \& \bibinfo{author}{Shaw, D.~E.}
  \bibinfo{title}{Parallel random numbers: as easy as 1, 2, 3}.
\newblock in \emph{\bibinfo{booktitle}{Proceedings of the International
  Conference for High Performance Computing, Networking, Storage and Analysis
  (SC '11)}} (\bibinfo{year}{2011}).


\bibitem{liu1989}
\bibinfo{author}{Liu, D.~C.} \& \bibinfo{author}{Nocedal, J.}
\newblock \bibinfo{title}{{On the limited memory BFGS method for large scale
  optimization}}.
\newblock \emph{\bibinfo{journal}{Math. Program.}}
  \textbf{\bibinfo{volume}{45}}, \bibinfo{pages}{503--528}
  (\bibinfo{year}{1989}).


\bibitem{armijo1966}
\bibinfo{author}{Armijo, L.}
\newblock \bibinfo{title}{{Minimization of functions having Lipschitz
  continuous first partial derivatives}}.
\newblock \emph{\bibinfo{journal}{Pac. J. Math.}}
  \textbf{\bibinfo{volume}{16}}, \bibinfo{pages}{1--3} (\bibinfo{year}{1966}).


\bibitem{byrd1994}
\bibinfo{author}{Byrd, R.~H.}, \bibinfo{author}{Nocedal, J.} \&
  \bibinfo{author}{Schnabel, R.~B.}
\newblock \bibinfo{title}{{Representations of quasi-Newton matrices and their
  use in limited memory methods}}.
\newblock \emph{\bibinfo{journal}{Math. Program.}}
  \textbf{\bibinfo{volume}{63}}, \bibinfo{pages}{129--156}
  (\bibinfo{year}{1994}).


\bibitem{shapovalov2011}
\bibinfo{author}{Shapovalov, M.~V.} \& \bibinfo{author}{Dunbrack, J.,
  Roland~L.}
\newblock \bibinfo{title}{A smoothed backbone-dependent rotamer library for
  proteins derived from adaptive kernel density estimates and regressions}.
\newblock \emph{\bibinfo{journal}{Structure}} \textbf{\bibinfo{volume}{19}},
  \bibinfo{pages}{844--858} (\bibinfo{year}{2011}).


\bibitem{PISCES}
\bibinfo{author}{Wang, G.} \& \bibinfo{author}{Dunbrack, J., Roland~L.}
\newblock \bibinfo{title}{{PISCES}: a protein sequence culling server}.
\newblock \emph{\bibinfo{journal}{Bioinformatics}}
  \textbf{\bibinfo{volume}{19}}, \bibinfo{pages}{1589--1591}
  (\bibinfo{year}{2003}).


\bibitem{conway2016}
\bibinfo{author}{Conway, P.} \& \bibinfo{author}{DiMaio, F.}
\newblock \bibinfo{title}{Improving hybrid statistical and physical forcefields
  through local structure enumeration}.
\newblock \emph{\bibinfo{journal}{Protein Sci.}} \textbf{\bibinfo{volume}{25}},
  \bibinfo{pages}{1525--1534} (\bibinfo{year}{2016}).


\bibitem{casf2016}
\bibinfo{author}{Su, M.} \emph{et~al.}
\newblock \bibinfo{title}{{Comparative assessment of scoring functions: the
  CASF-2016 update}}.
\newblock \emph{\bibinfo{journal}{J. Chem. Inf. Model.}}
  \textbf{\bibinfo{volume}{59}}, \bibinfo{pages}{895--913}
  (\bibinfo{year}{2019}).

\end{thebibliography}
\end{document}